\pdfoutput=1
\documentclass[conference]{IEEEtran}
\IEEEoverridecommandlockouts

\usepackage{algorithm}
\usepackage{algpseudocode}
\usepackage{amsmath,amssymb,amsfonts}
\usepackage{graphicx}
\usepackage{textcomp}
\def\BibTeX{{\rm B\kern-.05em{\sc i\kern-.025em b}\kern-.08em
    T\kern-.1667em\lower.7ex\hbox{E}\kern-.125emX}}

\usepackage[english]{babel}
\usepackage{subfigure}
\usepackage{epsfig}
\usepackage{multirow}
\usepackage{etoolbox}
\usepackage{color}

\usepackage{hyperref}       
\usepackage{url}            
\usepackage{booktabs}       
\usepackage{nicefrac}       
\usepackage{microtype}      
\usepackage{bm}
\usepackage{enumitem}

\newcommand{\method}[1]{\textsl{#1}} 
\newcommand{\task}[1]{\textit{#1}} 

\newcommand{\mat}[1]{\bm{#1}} 

\newcommand{\field}[1]{\mathbb{#1}}
\newcommand{\R}{\field{R}} 

\DeclareMathOperator*{\E}{\mathbb{E}}

\newcommand{\ehrgan}{\texttt{ehrGAN}}

\begin{document}

\title{Boosting Deep Learning Risk Prediction with Generative Adversarial Networks for Electronic Health Records}

\author{
\IEEEauthorblockN{
Zhengping Che\IEEEauthorrefmark{2}\IEEEauthorrefmark{1},
Yu Cheng\IEEEauthorrefmark{3}\IEEEauthorrefmark{1}\thanks{\IEEEauthorrefmark{1}Equal contributions.},
Shuangfei Zhai\IEEEauthorrefmark{4},
Zhaonan Sun\IEEEauthorrefmark{6},
Yan Liu\IEEEauthorrefmark{2}
}
\IEEEauthorblockA{\IEEEauthorrefmark{2}{University of Southern California},
Los Angeles, CA, USA 90089 \\
zche@usc.edu, yanliu.cs@usc.edu}
\IEEEauthorblockA{\IEEEauthorrefmark{3}{AI Foundations, IBM Thomas J. Watson Research Center},
Yorktown Heights, NY, USA 10598\\
chengyu@us.ibm.com}
\IEEEauthorblockA{\IEEEauthorrefmark{4}{Binghamton University, SUNY},
Binghamton, NY, USA 13902 \\
szhai2@binghamton.edu}
\IEEEauthorblockA{\IEEEauthorrefmark{6}{IBM Thomas J. Watson Research Center},
Yorktown Heights, NY, USA 10598\\
zsun@us.ibm.com}
}

\maketitle

\begin{abstract}
The rapid growth of Electronic Health Records (EHRs), as well as the accompanied opportunities in Data-Driven Healthcare (DDH), has been attracting widespread interests and attentions.
Recent progress in the design and applications of deep learning methods has shown promising results and is forcing massive changes in healthcare academia and industry, but most of these methods rely on massive labeled data.
In this work, we propose a general deep learning framework which is able to boost risk prediction performance with limited EHR data.
Our model takes a modified generative adversarial network namely \textit{\ehrgan}, which can provide plausible labeled EHR data by mimicking real patient records, to augment the training dataset in a semi-supervised learning manner.
We use this generative model together with a convolutional neural network (CNN) based prediction model to improve the onset prediction performance.
Experiments on two real healthcare datasets demonstrate that our proposed framework produces realistic data samples and achieves significant improvements on classification tasks with the generated data over several stat-of-the-art baselines. 
\end{abstract}

\begin{IEEEkeywords}
electronic health record; generative adversarial network; health care; deep learning;
\end{IEEEkeywords}

\section{Introduction}
\label{sec:intro}
The worldwide exponential surge in volume, detail, and availability of \textit{Electronic Health Records (EHRs)} promises to usher in the era of personalized medicine, enhancing each stage of the healthcare chain from providers to patients. This field of research and applications, commonly mentioned as \textit{Data-Driven Healthcare (DDH)}~\cite{madsen2014data}, has been under rapid development and attracted many researchers and institutions to utilize state-of-the-art machine learning and statistical models on a broad set of clinical tasks which are difficult or even impossible to solve with traditional methods~\cite{wu2010prediction, raghupathi2014big,kleinberg2011review}.
Among these frontier models, deep learning tends to be the most exciting and promising solution to those difficult and important tasks.

Recent success and development in deep learning is revolutionizing many domains such as computer vision~\cite{he2016deep,karpathy2014large}, natural language processing~\cite{cho2014learning,sutskever2014sequence}, and healthcare, with notable innovations and applicable solutions.
A series of excellent work have been conducted in seek of novel deep learning solutions in different healthcare applications including but not limited to computational phenotyping~\cite{lasko2013computational,Che:2015:DCP:2783258.2783365}, risk prediction~\cite{DBLP:conf/sdm/ChengWZH16, citeulike:14040136}, medical imaging analysis~\cite{xu2014deep,esteva2017dermatologist}, and clinical natural language processing~\cite{muneeb2015evaluating}.
These works have made us closest ever towards the ultimate goal of improving health quality, reducing cost, and most importantly saving lives.

While existing achievements on deep learning models for healthcare are encouraging, the peculiar properties of EHRs, such as heterogeneity, longitudinal irregularity, inherent noises, and incomplete nature, make it extremely difficult to apply most existing mature models to healthcare compared with other well-developed domains with clean data.
Properly-designed deep neural networks have the prospect of handling these issues if equipped with massive data, but the amount of clinical data, especially with accurate labels and for rare diseases and conditions, is somewhat limited and far from most models' requirements~\cite{koh2011data}. This comes from the following reasons:
The diagnosis and patient labeling process highly relies on experienced human experts and is usually very time-consuming;
Getting detailed results of lab tests and other medical features, though has become more feasible with modern facilities than ever, are still quite costly;
Not to mention the privacy issues and regulations which makes it even harder to collect and keep enough medical data with desired details.
These unique challenges lying in healthcare domain prevent existing deep learning models from exerting their strength with enough available and high-quality labeled data.

One way to overcome the challenges arising from limited data in machine learning field is \textit{semi-supervised learning (SSL)}~\cite{chapelle2009semi}.
Semi-supervised learning is a class of techniques that makes use of unlabeled or augmented data together with a relatively small set of labeled data to get better performance. Though some previous work utilized semi-supervised learning methods on EHR data~\cite{zhang2015semi}, most of them focus on clinical text data~\cite{wongchaisuwat2016semi,garla2013semi}, and only limited work attempt to perform semi-supervised learning method on structured quantitative EHR data~\cite{beaulieu2016semi}.

Generative model is also considered as a promising solution. As one type of semi-supervised learning algorithms, it aims at learning the joint probability distribution over observations and labels from the training data, and can be further used for downstream algorithms and applications such as data modeling~\cite{taylor2007modeling}, classifier and predictor training~\cite{ng2002discriminative}, and data augmentations~\cite{zhu2014gibbs}.
Though generative model approaches have been well explored for years, deep generative models haven't caught enough attentions due to its complexity and computation issues until the recent development of \textit{generative adversarial network (GAN)}~\cite{gan}. GAN simultaneously trains a deep generative model and a deep discriminative model, which captures the data distribution and distinguishes generated data from original data respectively, as a mini-max game.
GANs have been mainly used on image, video and text data to learn useful features with better understandings~\cite{radford2015unsupervised,mathieu2016disentangling} or sample data for specific demand~\cite{wang2016learning, im2016generating}.
However, few GANs have been applied for generating sequential or time series EHR data, where large amount of reliable data, either from real dataset or augmentations, are in great demand for powerful predictive models.

In this paper, we investigate and propose general deep learning solutions to the challenges on high dimensional temporal EHR data with limited labels.
We propose a generative model, {\ehrgan}, for EHR data via adversarial training, which is able to generate convincing and useful samples similar to realistic patient records in existing data.
We further propose a semi-supervised learning framework which achieves boosted risk prediction performance by utilizing the augmented data and representations from the proposed generative models.
We conduct experiments on two real clinical tasks and demonstrate the efficacy of both the generative model and prediction framework.

%
%

\section{Related Work}
\label{sec:review}
In this section we briefly review existing works which are closely related to the our proposed method in this paper from two areas. The first one is recent works on exploiting deep learning methods to applications on Electronic Health Records. The other is on the recent research of adversarial training and generative adversarial networks.

\subsection{Deep Learning for Healthcare Applications}
As deep learning has achieved great success recently, researchers have begun attempting to apply neural network based methods to EHR to utilize the ability of deep networks to learn complex patterns from data.
Previous studies, such as phenotype learning~\cite{Che:2015:DCP:2783258.2783365} and representation learning~\cite{citeulike:14040136}, formed a multi-layer perception (MLP) architecture and applied it to EHRs.
Considering the natural temporarily in EHR data, Recurrent neural networks (RNNs) are used for sequence prediction with regular times series of real-valued variables collected from intensive care unit patients and interpretable prediction on the diagnosis code given medication, lab test and disease information~\cite{Choi:2016:MRL:2939672.2939823}. Deep state space model is also designed for phenotype learning and handled the data irregularity issue~\cite{DBLP:conf/ichi/GhoshCS16}.
Several other works~\cite{DBLP:journals/corr/RazavianS15,DBLP:conf/sdm/ChengWZH16,DBLP:journals/corr/CheCSL17} exploited convolutional neural networks (CNNs) as well, to capture local temporal dependency of data in risk prediction or other related tasks.

All of these models are purely or mainly in the supervised learning manner and use fully labeled data.
However, supervised information is limited for many EHR applications, since it requires expense human efforts in labeling or scoring the patient records so as to analyze them in a model. Thus unsupervised and semi-supervised learning schema is in great demand to be designed.
There are some other works~\cite{Choi:2016:MRL:2939672.2939823,DBLP:journals/corr/CheCSL17} trying to learn the medical feature and concept embedding representations with unsupervised method and the learned representations  incorporate both co-occurrence information and visit sequence information of the EHR data. However, to our best of knowledge, none of existing work attempts to build a semi-supervised deep learning model for applications with time series EHR data, and our work is the first of its kind.

In health informatics domain, there are some works targeting semi-supervised problems in both deep and non-deep settings, but most of them focus on clinical natural language processing problems, including learning from structured quantitative EHR data~\cite{beaulieu2016semi}, building graph-based model for clinical text classification~\cite{garla2013semi}, and handling question-answering task in healthcare forum data~\cite{wongchaisuwat2016semi}.
They are not directly related to our work since we consider different data types or tasks.

\subsection{Adversarial Learning and Generative Adversarial Networks}
The idea of adversarial learning is to include a set of machines which learn together by pursuing competing goals.
In generative adversarial networks (GANs)~\cite{gan}, a generator function (usually formed as a deep neural network) learns to synthesize samples that best resemble some dataset, while a discriminator function (also usually a deep neural network) learns to distinguish between samples drawn from the dataset and samples synthesized by the generator.
There are lots of deep learning works out on GANs recently, and some of them have emerged as promising frameworks for unsupervised learning.
For instance, the generators are able to produce images of unprecedented visual quality~\cite{li2016precomputed}, while the discriminators learn features with rich semantics that can benefit other learning paradigms such as semi-supervised learning and transfer learning~\cite{NIPS2016_6544,DBLP:journals/corr/SalimansGZCRC16}.

The semi-supervised learning frameworks with GAN are used to solve classification tasks and learn a generative model simultaneously. The representations learned by the discriminator, the classifications from the semi-supervised classifier, and the sampled data from the generator improve each other.
There are several works to build theoretical semi-supervised learning frameworks with GANs and apply them to the classification task.
Generally speaking, existing methods include feature augmentation~\cite{DBLP:journals/corr/Springenberg15}, virtual adversarial training~\cite{DBLP:journals/corr/MiyatoDG16}, and joint training~\cite{DBLP:journals/corr/SalimansGZCRC16,DBLP:journals/corr/PapernotAEGT16}.
Our proposed semi-supervised paradigm belongs to data augmentation methods.

It is noting that all these related models are only applied to and designed for vision or natural language processing (NLP) domains. To extend existing GAN framework to EHR data is not straightforward. Moreover, to facilitate GANs with semi-supervised learning for onset prediction is also difficult. These two unsolved challenges are well addressed in our proposed framework. 

\section{The Proposed Method}
\label{method}
In this section, we first introduce the basic deep learning risk prediction model used as a strong baseline as well as a component in the proposed framework. Then we describe \ehrgan, a modified generative adversarial network which is specifically designed to be applied on EHR data. Finally, we present the data augmented semi-supervised learning schema which is able to perform boosted onset predictions.

\subsection{Basic Deep Prediction Model}
\label{sec:method-basicmodel}
The basic model used in the paper is a convolutional neural network (CNN) model with 1D convolutional layer over the temporal dimension and max over-time pooling layer, which was used in previous work~\cite{DBLP:journals/corr/RazavianS15,DBLP:conf/sdm/ChengWZH16,DBLP:journals/corr/CheCSL17}.
The input to the model is the EHR records of patient $p$, which is represented as a temporal embedding matrix $\mat{x}^p \in \R^{T_p \times M}$, where $T_p$, which can be different among patients, is the number of medical events in patient $p$'s record, and $M$ is the dimension of the learned embedding.
The rows of $\mat{x}^p$ are the embedding vectors for the medical events, arranged in the order of time of occurrence.
The embedding for medical events is trained by Word2vec model~\cite{mikolov2013distributed} on the same corpus of EHR data.
We apply convolutional operation only over the temporal dimension but not over embedding dimension.
We us a combination of filters with different lengths to capture temporal dependencies in multiple levels, and our preliminary experiments validated the performance improvement from such strategy.
After the convolutional step, we apply a max-pooling operation along the temporal dimension to keep the most important features across the time. This temporal pooling converts the inputs with different temporal lengths into a fixed length output vector. Finally a fully connected soft-max layer is used to produce prediction probabilities.
This CNN-based deep prediction model described above is shown to be the most competitive baseline among all other tested baselines in our experiments and serves as the basic prediction component in our proposed work.

\subsection{{\ehrgan}: Modified GAN Model for EHR Data}
\label{sec:method-ehrgan}
The original GAN \cite{gan} is trained by solving the following mini-max game in the form of
\begin{equation*}
\min_G \max_D
\E_{\mat{x} \sim p_{data}(\mat{x})}\left[\log D(\mat{x})\right] +
\E_{\mat{z} \sim p_{\mat{z}}(\mat{z})}\left[\log\left(1 - D(G(\mat{z}))\right)\right]
\end{equation*}
where $p_{data}(\mat{x})$ is the true data distribution;
$D(\mat{x})$ is the discriminator that takes a sample as the input and outputs a scalar between $[0,1]$ as the probability of the sample drawing from real dataset;
$G(\mat{z})$ is the generator that maps a noise variable $\mat{z} \in \R^d$ drawn from a given distribution $p_{\mat{z}}(\mat{z})$ back to the input space.
The training procedure consists of two loops optimizing $G$ and $D$ iteratively. After the mini-max game reaches its Nash equilibrium~\cite{nash1951non}, $G$ defines an implicit distribution $p_g(\mat{x})$ that recovers the data distribution, i.e., $p_g(\mat{x}) = p_{data}(\mat{x})$.

\begin{figure}
\centering
\subfigure[\label{fig:structure}The structure of {\ehrgan} model. $M$ and $\tilde{M}$ are the models in discriminator ($D$) and generator ($G$). $s$ and $\tilde{s}$ represent real and synthetic samples.
]{
\includegraphics[width=0.45\columnwidth]{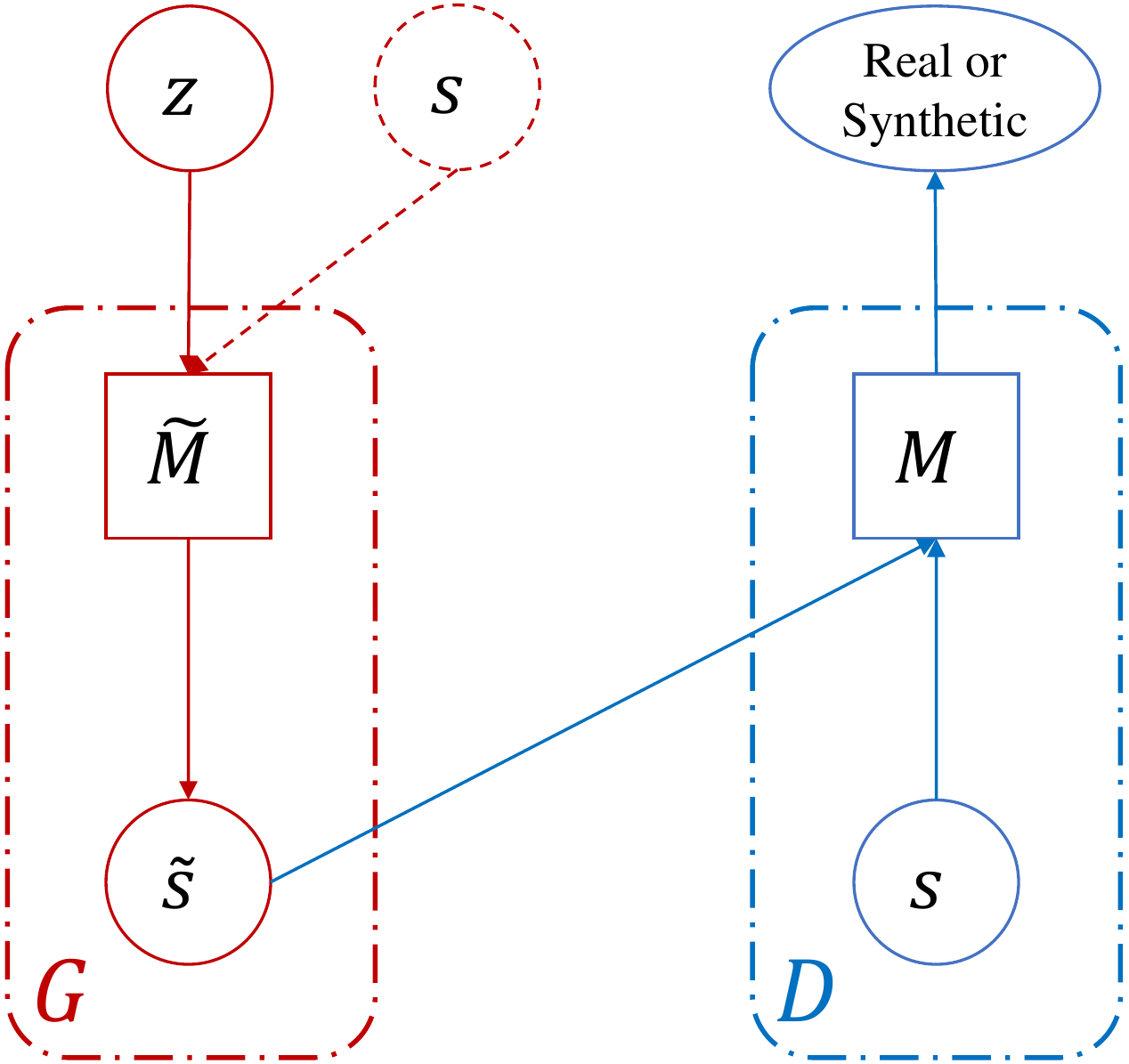}
}
\hfill
\subfigure[\label{fig:vcd}The structure of the generator in {\ehrgan}. $\mat{z}$ and $\mat{m}$ are drawn randomly. $\mat{\tilde{x}} $ is the generated synthetic sample based on the real sample $\mat{x}.$]{
\includegraphics[width=0.43\columnwidth]{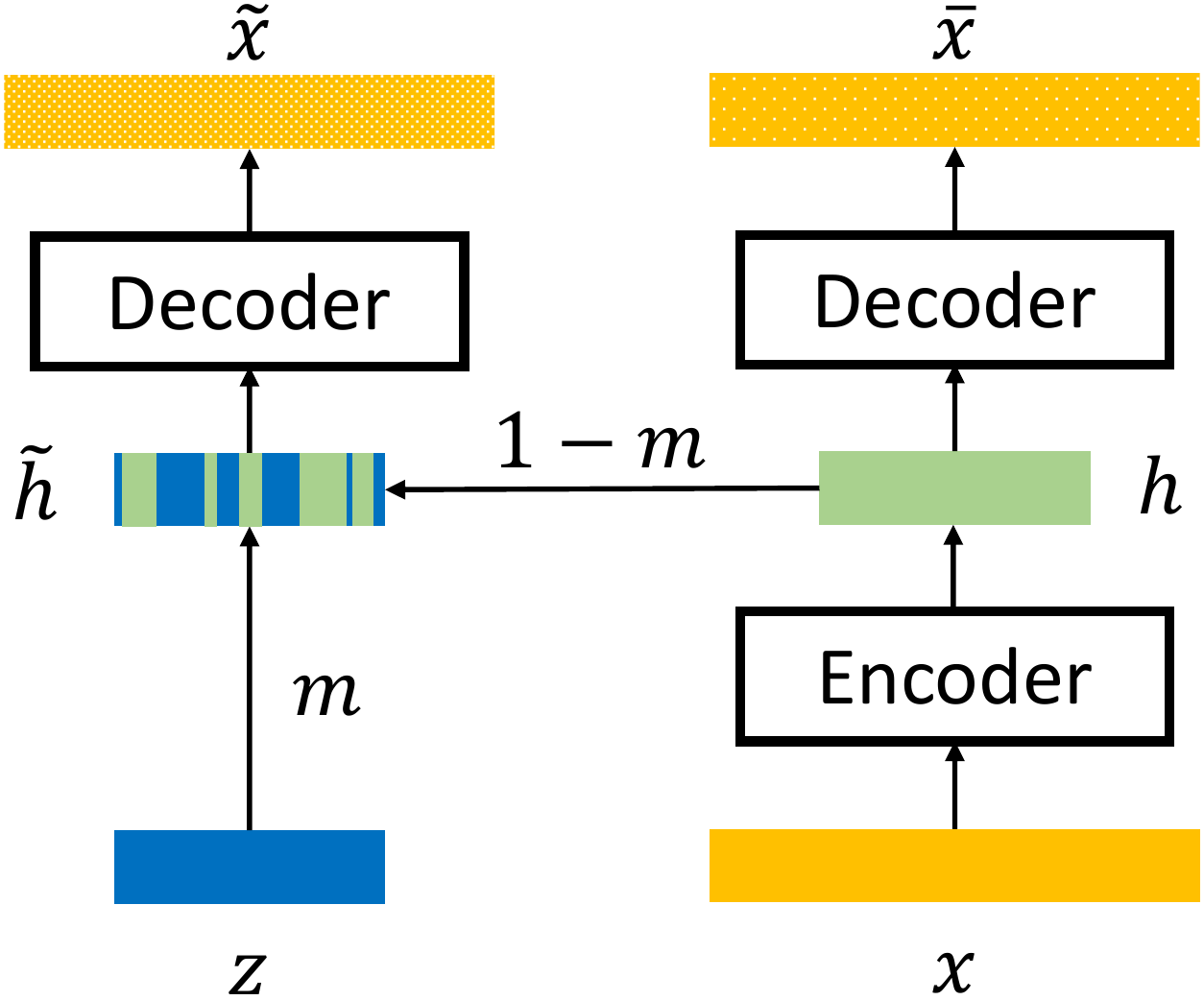}
}
\caption{Illustration of the proposed {\ehrgan} model.}
\end{figure}

Generally, both $D$ and $G$ are parameterized as deep neural networks.
In the context of EHR data, similar to the basic prediction models, our choice of $G$ and $D$ falls into the family of 1D convolutional neural networks (CNNs) and 1D deconvolutional neural networks (DCNNs).
The overview of the model is shown in Figure~\ref{fig:structure}. In the following parts, we will discuss model details in terms of the design of discriminator and generator, and some specific training techniques.


\subsubsection{Discriminator}
We adopt the structure of the basic prediction model to the discriminator, due to its simplicity and excellent classification performance. We replace the top prediction layer by a single sigmoid unit to output the probability of  the input data being drawn from the real dataset.

\subsubsection{Generator}
The goal of the generator in GAN is to translate a latent vector $\mat{z}$ into the synthetic sample $\mat{\tilde{x}}$. The generator is encoded by a de-convolutional neural network with two consecutive fully connected layers, the latter of which is reshaped and followed by two de-convolutional layers to perform upsampling convolution. 
Empirically, this generator is able to generate good samples. However, this version of generator can not be directly used in semi-supervised learning setting as the model is trained only to differentiate real or synthetic data instead of the classes.
To solve this problem, we introduce a variational version of the generator, which also provides some new understandings of GANs.

\paragraph{Generator with variational contrastive divergence}
The design of the variational generator is based on the recently proposed variational contrastive divergence (VCD)~\cite{vgan}.
Instead of directly learning a generator distribution defined by $G(\mat{z})$, we learn a transition distribution of the form $p(\mat{\tilde{x}}|\mat{x})$ for the generated sample $\mat{\tilde{x}}$, with $\mat{x} \sim p_{data}(\mat{x})$.
The marginal distribution of of the generator is then given by $p_g(\mat{\tilde{x}}) = \E_{\mat{x} \sim p_{data}(\mat{x})}p(\mat{\tilde{x}}|\mat{x})$.
Intuitively, the transition distribution $p(\mat{\tilde{x}}|\mat{x})$ encodes a generating process. In this process, based on an example drawn from the training data distribution, a neighboring sample $\mat{\tilde{x}}$ is generated.
To be more specific, the generator is equipped with encoder-decoder CNN networks.
For each real sample $\mat{x}$, we can get the representation $\mat{h}$ from encoder, and the reconstruction $\mat{\bar{x}}$ by feeding $\mat{h}$ into decoder.
$\mat{h}$ can be mixed with a random noise vector $\mat{z}$ of the same dimensionality by a random binary mask vector $\mat{m}$ to obtain $\mat{\tilde{h}} = \mat{m} * \mat{z} + \left(\mat{1} - \mat{m}\right) * \mat{h}$, where $*$ represents element-wise multiplication. The synthetic sample $\mat{\tilde{x}}$ can be obtained by feeding $\mat{\tilde{h}}$ to the same decoder.
An illustration of this VCD-based generator is shown in Figure~\ref{fig:vcd}.
The generator attempts to minimize the objective as
\begin{equation}
\label{eq:ae}
\E_{\mat{x} \sim p_{data}(\mat{x})} \left[
\rho \cdot \E_{\mat{\tilde{x}} \sim p_g(\mat{\tilde{x}}|\mat{x})} \left[ - \log D(\mat{\tilde{x}})\right]
+ (1 - \rho) \cdot {\|\mat{\bar{x}} - \mat{x}\|}_2^2
\right]
\end{equation}
where $D$ is the discriminator function and the hyperparameter $\rho$ controls how close the synthetic sample should be to the corresponding real sample.
The usage of VCD-based {\ehrgan} brings two benefits.
First, while original GANs are known to have \textit{mode collapsing} issues, i.e., $G$ is encouraged to generate only a few modes, {\ehrgan} eliminates \textit{mode collapsing} issue by its design, as the diversity of the generated samples inherently approximates that of the training data.
Second and more importantly, the learned transition distribution $p(\mat{\tilde{x}}|\mat{x})$ contains rich structures of the data manifold around training examples $\mat{x}$, which can be quite useful when incorporating with our semi-supervised learning framework to obtain effective classification models.



\subsubsection{Training techniques}
We train the proposed {\ehrgan} by optimizing the generator and discriminator iteratively with stochastic gradient descent (SGD). The training procedure (shown in Algorithm~\ref{alg:vcd}) is similar to that of standard GANs.
We take several techniques to stabilize the training of GANs similar to those in \cite{mathieu2016disentangling,DBLP:journals/corr/SalimansGZCRC16}, and relieve the training instability and sensitivity to hyper-parameters.
Firstly, we switch the order of discriminator and generator training, and perform $k=5$ optimization steps for the generator for every one steps for the discriminators.
Secondly, we add an $l_2$-norm regularizer in the cost function of discriminator.
Finally, batch normalization and label smoothing techniques are used.

\begin{algorithm}[h]
\caption{The optimization procedure of {\ehrgan}}
\label{alg:vcd}
\begin{algorithmic}[1]
\small
\For{enough iterations until convergence}
\For{$k$ inner steps}
\State sample $N$ noise variables $\{\mat{z}^1, \dots, \mat{z}^N\}$, and $N$ binary mask vectors $\{\mat{m}^1, \dots, \mat{m}^N\}$;
\State update \textit{generator $G$} by one step gradient \textit{ascent} of \par
$ \qquad
\frac{1}{N}\sum_{i=1}^N \log D({G(\mat{z}^i, \mat{m}^i)})
$
\EndFor
\State sample $N$ training data $\{\mat{x}^1, \dots, \mat{x}^N\}$, $N$ noise variables $\{\mat{z}^1, \dots, \mat{z}^N\}$, and $N$ binary mask vectors $\{\mat{m}^1, \dots, \mat{m}^N\}$;
\State update \textit{discriminator $D$} with one step gradient \textit{descent} of \par
$
-\frac{1}{N}\sum_{i=1}^N \log D({\mat{x}^i}) - \frac{1}{N}\sum_{i=1}^N \log\left(1 - D({G(\mat{z}^i, \mat{m}^i)})\right)
$
\EndFor
\end{algorithmic}
\end{algorithm}

\subsection{Semi-supervised Learning with GANs}
\label{method-ssl-gan}
We next introduce our method of conducting semi-supervised learning (SSL) with a learned {\ehrgan}, in a way which is similar to our previous model for images~\cite{vgan}. The basic idea is to use the learned transition distribution to perform data augmentation. To be concrete, within the SSL setting we minimize the follow loss function:
\begin{equation}
\label{eq:semi}
\frac{1}{N}\sum_{i=1}^N\mathcal{L}(\mat{x}^i, \mat{y}^i) + \mu \cdot \frac{1}{N}\sum_{i=1}^N \E_{\mat{\tilde{x}}^i \sim p(\mat{\tilde{x}} | \mat{x}^i)}\mathcal{L}(\mat{\tilde{x}}^i, \mat{y}^i)
\end{equation}
where $\mathcal{L}$ refers to the binary crossentropy loss on each data sample, and $\mu$ leverages the ratio of the numbers of training data and augmented data from GANs. In other words, this model assumes that a well trained generator with distribution $p(\mat{\tilde{x}}|\mat{x})$ should be able to generate samples that are likely to align within the same class of $\mat{x}$, which can in turn provide valuable information to the classifier as additional training data. This method is called \textbf{\method{SSL-GAN}} (Semi-supervised learning with a learned {\ehrgan}) in this paper.

\section{Experimental Results} \label{experiment}
In this section we apply our models to two real clinical datasets extracted from heart failure and diabetes cohorts. It is a particularly interesting to investigate how well GANs can generate EHRs samples as the real ones.
Also, understanding how the proposed method can boost the performance of onset prediction is crucial for many healthcare applications. We start this section by introducing the datasets and experimental settings, and provide the evaluation analysis, followed by the discussions on the selections of parameters.

\subsection{Datasets and Settings}
The datasets came from a real-world longitudinal Electronic Health Record database of $218,680$ patients and $14,969,489$ observations of $14,690$ unique medical events, between the year 2011 to 2015 from a health insurance company. In these datasets, a set of diseases related ICD-9 codes were recorded to indicate medical conditions as well as drug prescriptions. 
We identify two following cohorts and predict whether a patient is from case or control group as a binary classification task. The labels of both case and control groups are identified by domain experts according to ICD-9 codes. 
\begin{itemize}
\item Congestive heart failure (\task{Heart Failure}), which contains $3,357$ confirmed patients in case group and $6,714$ patients in control group;
\item Diabetes (\task{Diabetes}), with $2,248$ patients in case group and $4,496$ patients in control group.
\end{itemize}
We import ICD-9 diagnosis and medications as the input features, eliminate those which show less than 5 times in this dataset, and get $8,627$ unique medical features.  We segmented the time dimension into disjoint 90-day windows and combined all the observations within each window. 
We split datasets into training, validation and test with ratio 7:1:2, and limit the length of each record sequence between $50$ and $250$ and form it to the embedding matrix. All sequences are 0-padded to match the longest sequence.
The embedding is trained by Word2vec~\cite{mikolov2013distributed} on the entire dataset with dimension of $200$. 
The {\ehrgan} is trained on only the training subset. 
For the CNN discriminator, we employ filters of sizes $\{3,4,5\}$ with $100$ feature maps. For the generator, the dimension of the latent variable $\mat{z}$ is $100$. It is first projected and reshaped by the generator and up-sampled by two one-dimensional CNN layers with filers size $100$ and $3$. The output of the generator is an embedding matrix with size $200\times 150$. These hyperparameters are selected based on preliminary experimental results.
To generate samples with different length, we paddle a special embedding mark at the end of each training record. The masks $\mat{m}$ in the VCD-based generator is uniformly sampled with probability $0.5$.
The Adam algorithm~\cite{kingma2014adam} with learning rate $0.001$ for both discriminator and generator is utilized for optimization. Gradients are clipped if the norm of the parameter vector exceeds $5$~\cite{sutskever2014sequence}. After we get the generated data, we can map it into EHR record by finding the nearest-neighbor with cosine distance for each feature.
The selection of optimal values for hyper-parameters $\mu$ and $\rho$ will be discussed later in Section~\ref{sec:exp-params}.

\begin{table}[hbt]
\small
\centering
\caption{Prediction performance comparison.}
\label{tab:main}
\begin{tabular}{lcccc}
\toprule[0.12em]
 & \multicolumn{2}{c}{\task{Heart Failure}} & \multicolumn{2}{c}{\task{Diabetes}} \\ \cmidrule{2-5}
 &  \textbf{Accuracy} & \textbf{AUROC} & \textbf{Accuracy} & \textbf{AUROC} \\ \midrule \midrule
\textbf{\method{CNN}} & $\mathbf{0.8630}$ & $\mathbf{0.9329}$ & $\mathbf{0.9644}$ & $\mathbf{0.9789}$ \\ \midrule
\method{GRU} & $0.8578$ & $0.9129$ & $0.9304$ & $0.9659$ \\ \midrule
\method{LSTM} & $0.8511$ & $0.9103$ & $0.9448$ & $0.9553$ \\ \midrule
\method{LR} & $0.8494$ & $0.9052$ & $0.9066$ & $0.9681$ \\ \midrule
\method{SVM} & $0.8443$ & $0.9017$ & $0.8944$ & $0.9462$ \\ \midrule
\method{RF} & $0.8571$ & $0.9225$ & $0.9476$ & $0.9658$ \\
\bottomrule[0.12em]
\end{tabular}
\vspace{-0.1in}
\end{table}

\subsection{Risk Prediction Comparison on Basic Models} \label{sec:exp-pred}
First, we show the performance of our basic predict model (\method{CNN}), which explores the CNN model with pre-trained medical feature embedding and is a strong baseline even before boosted.
We compare it with logistic regression (\method{LR}), linear support vector machine (\method{SVM}), random forest (\method{RF}) and two other deep models, recurrent neural network models using gated recurrent units (\method{GRU}~\cite{cho2014properties}) and long short-term memory (\method{LSTM} \cite{journals/corr/LiptonKEW15}). For \method{LR}, \method{SVM} and \method{RF}, we use the same setting as mentioned in previous work~\cite{DBLP:journals/corr/CheCSL17}.
We follow the similar settings from existing work~\cite{journals/corr/LiptonKEW15} for \method{GRU} and \method{LSTM}.

Table~\ref{tab:main} shows the classification accuracy and AUROC (area under receiver operating characteristic curve) of all basic baseline models on the two prediction tasks.
\method{CNN} is among the best methods in \task{Heart Failure} task and significantly outperforms baselines in \task{Diabetes} task. The performance improvement mainly comes from the learned embeddings in heart failure task, and from CNN model structures in diabetes task. The other two deep models \method{GRU} and \method{LSTM} work well but can not beat \method{CNN}.

\subsection{Analysis of Generated Data}
\begin{figure}[tb]
\centering
\subfigure[Heart Failure]{
\includegraphics[width=0.45\linewidth]{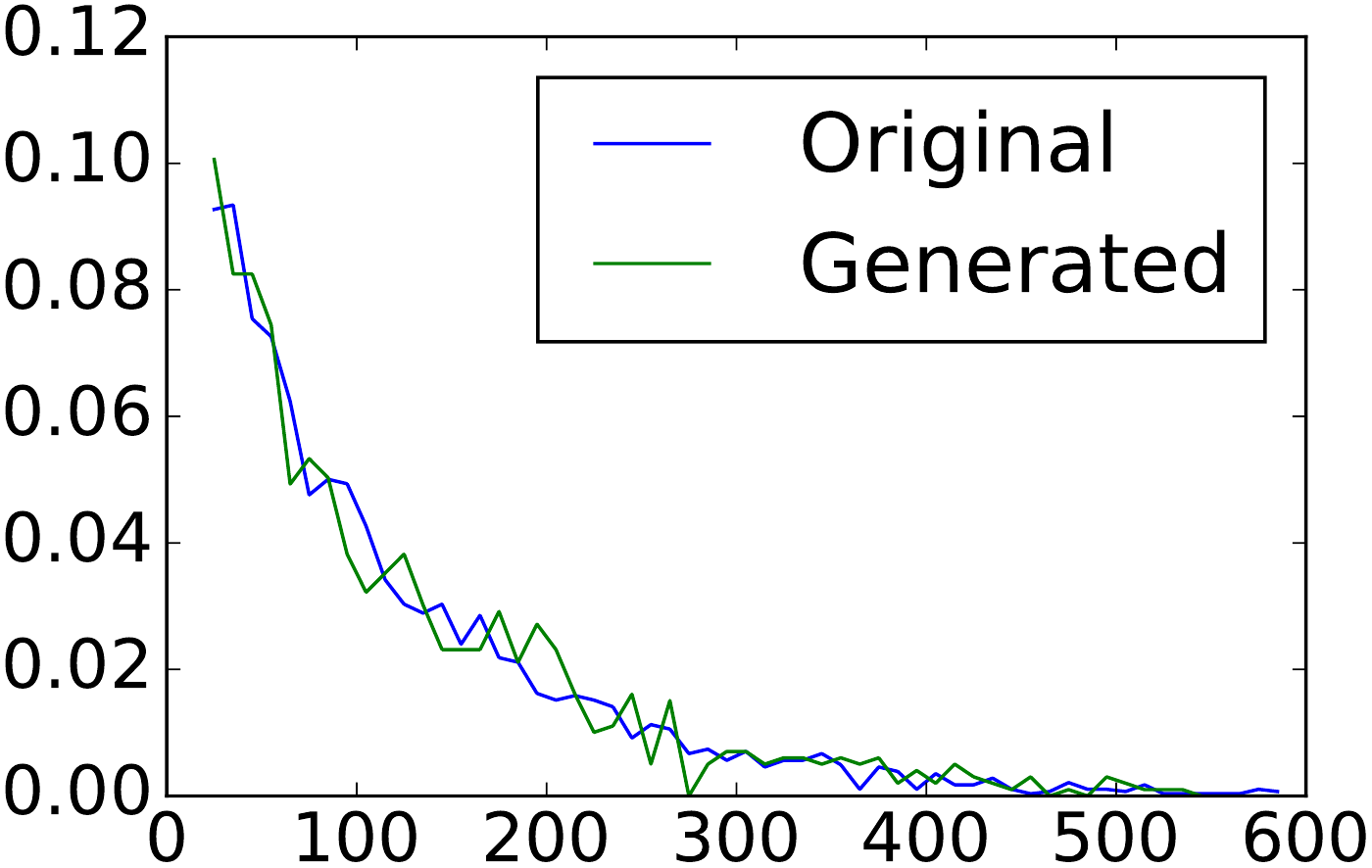}
}
\subfigure[Diabetes]{
\includegraphics[width=0.45\linewidth]{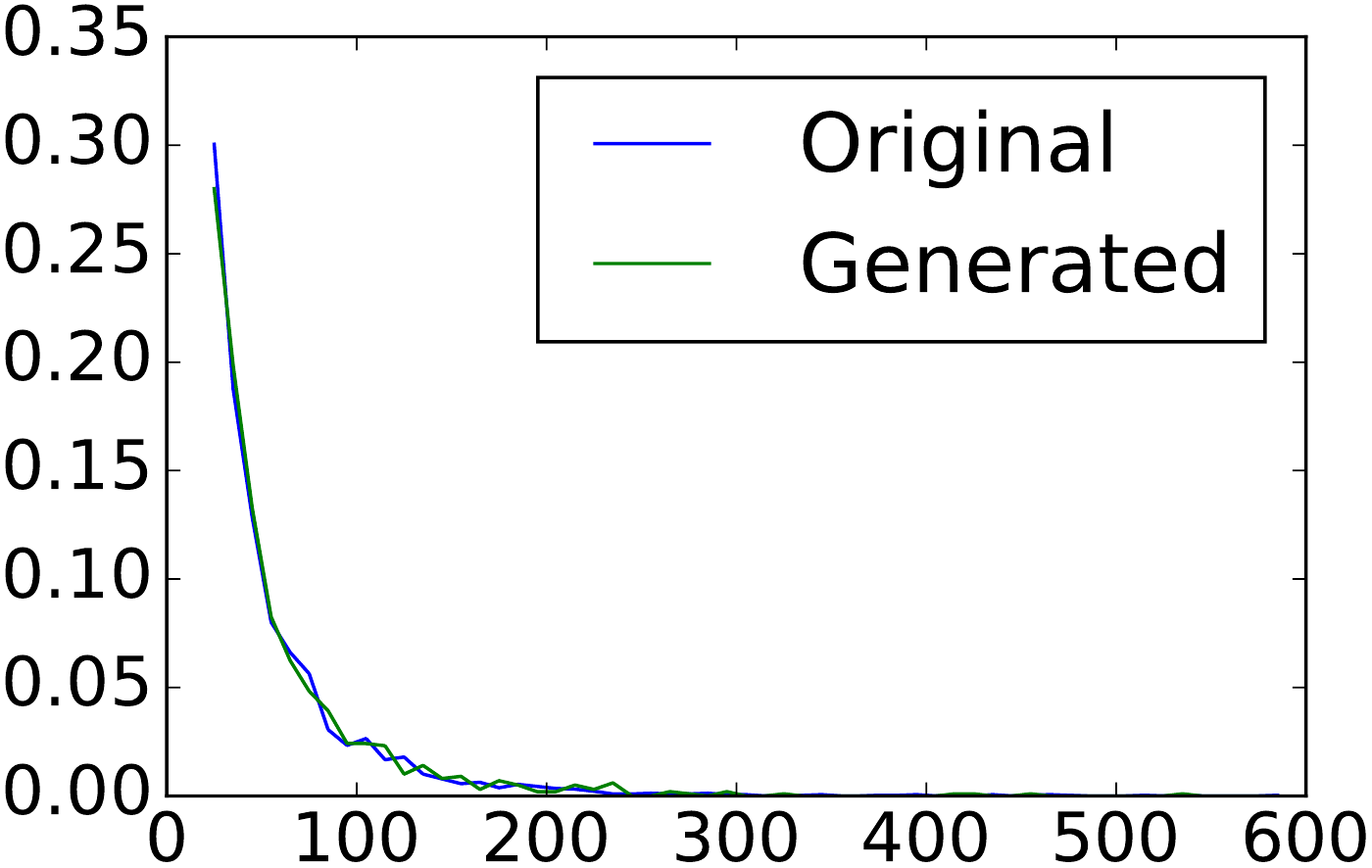}
}
\caption{The length distributions of original and generated datasets. x-axis: data length; y-axis: probability.}
\label{fig:gan-data-length}
\end{figure}

\begin{figure}[tb]
\centering
\subfigure[Heart Failure]{
\includegraphics[width=0.45\linewidth]{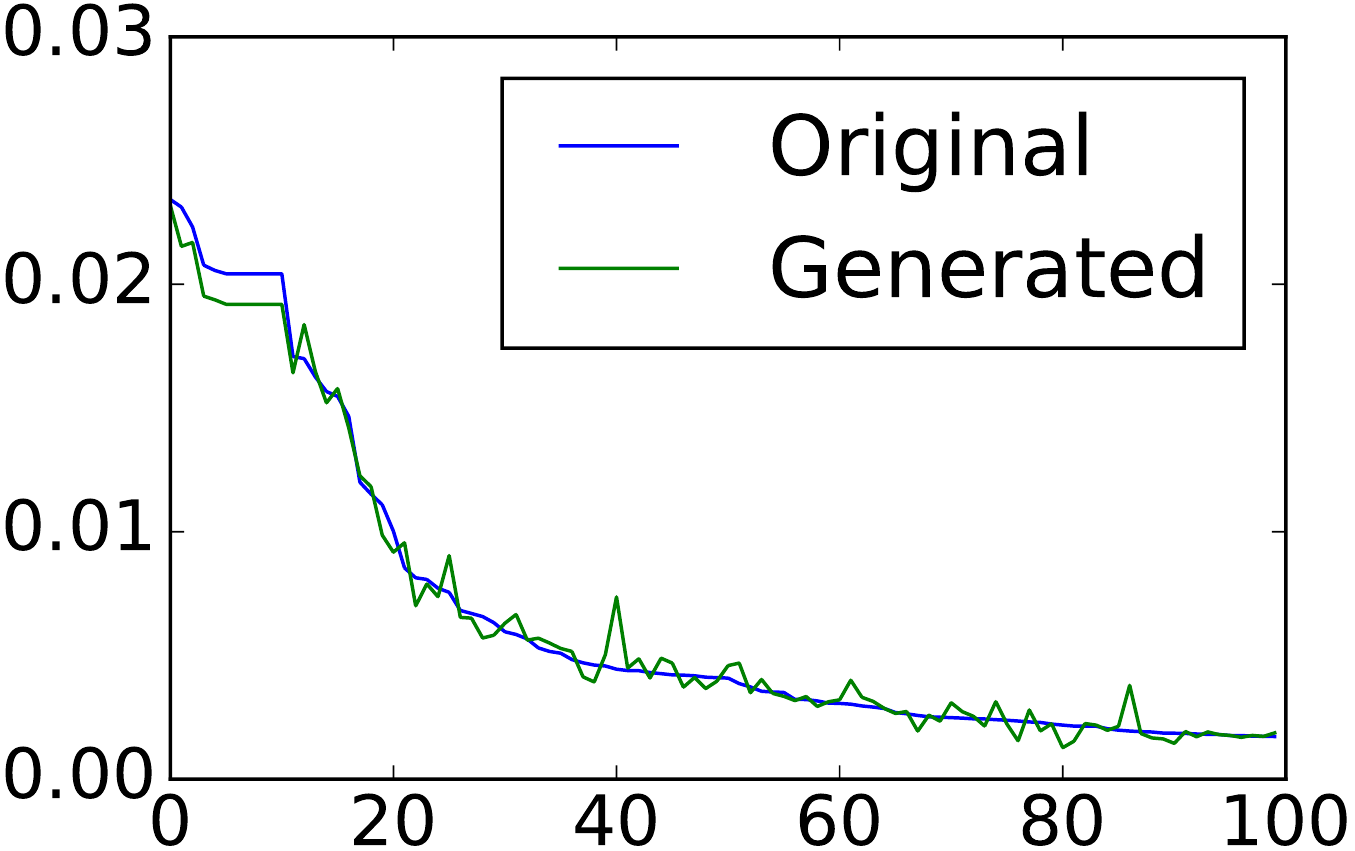}
}
\subfigure[Diabetes]{
\includegraphics[width=0.45\linewidth]{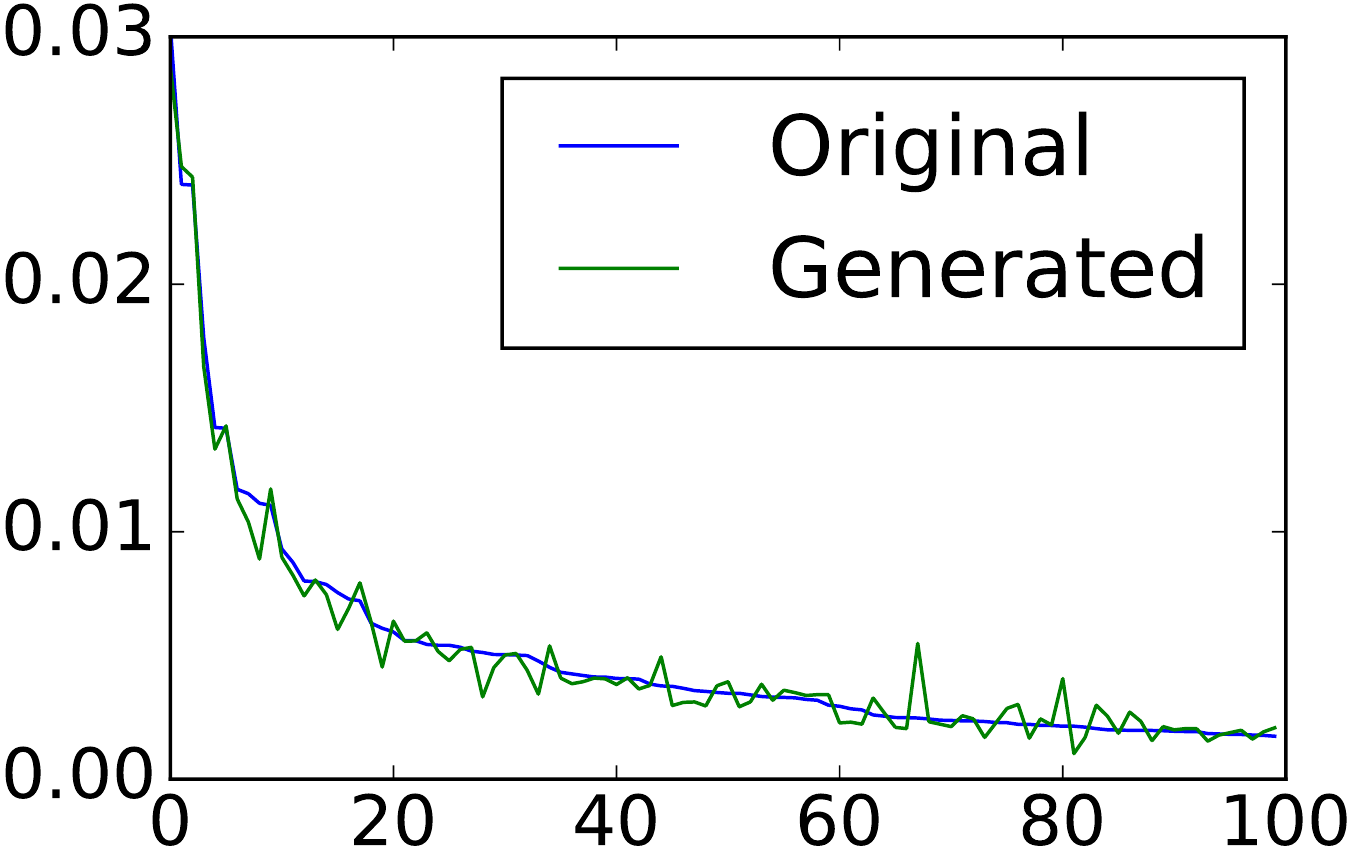}
}
\caption{The frequency of top 100 features in original and generated datasets. x-axis: features sorted by frequency in the original dataset. y-axis: frequency.}
\label{fig:gan-data-uni-dist}
\end{figure}

\begin{figure}[htb]
\centering
\subfigure[Heart Failure]{
\includegraphics[width=0.22\linewidth]{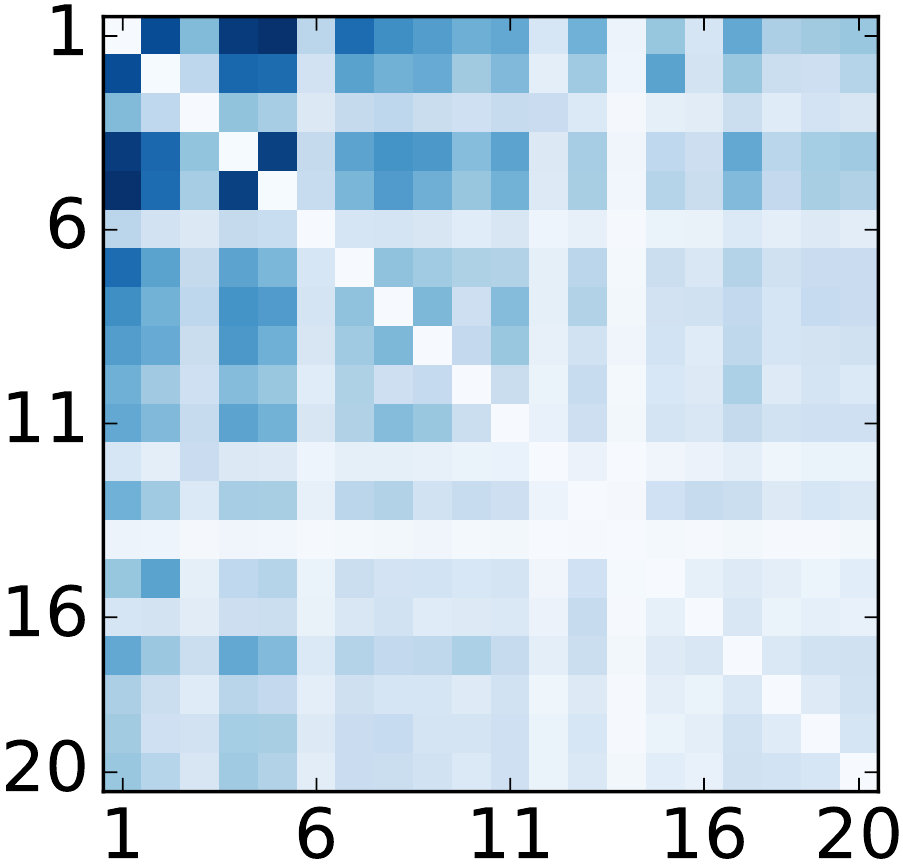}
\hfill
\includegraphics[width=0.22\linewidth]{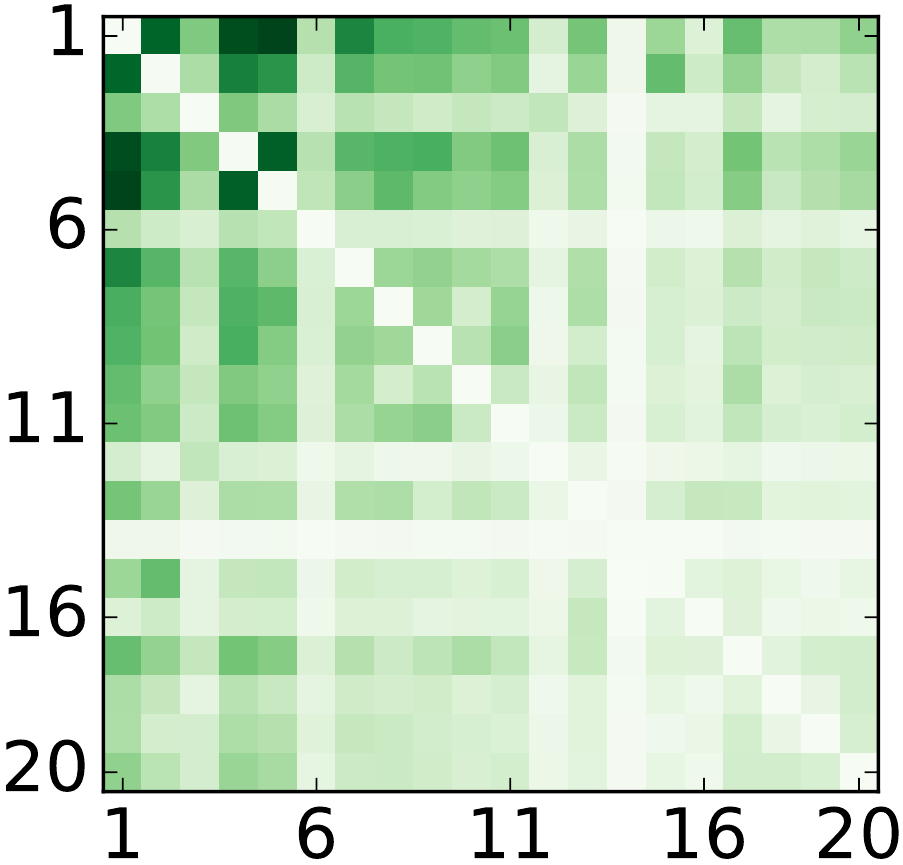}
}
\subfigure[Diabetes]{
\includegraphics[width=0.22\linewidth]{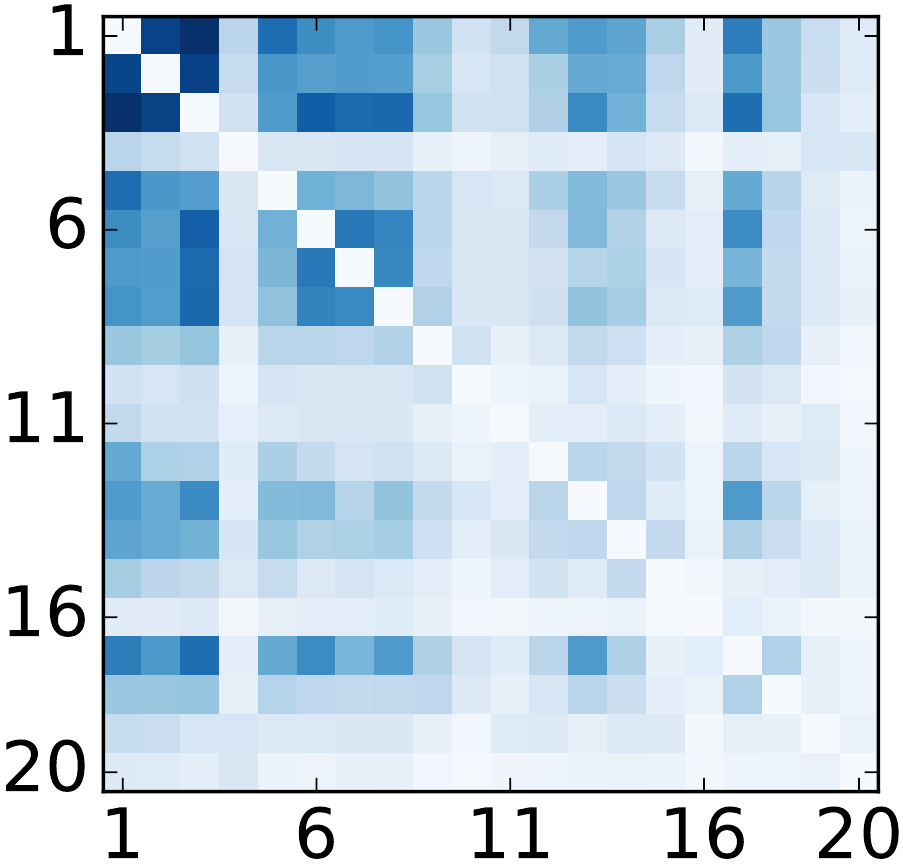}
\hfill
\includegraphics[width=0.22\linewidth]{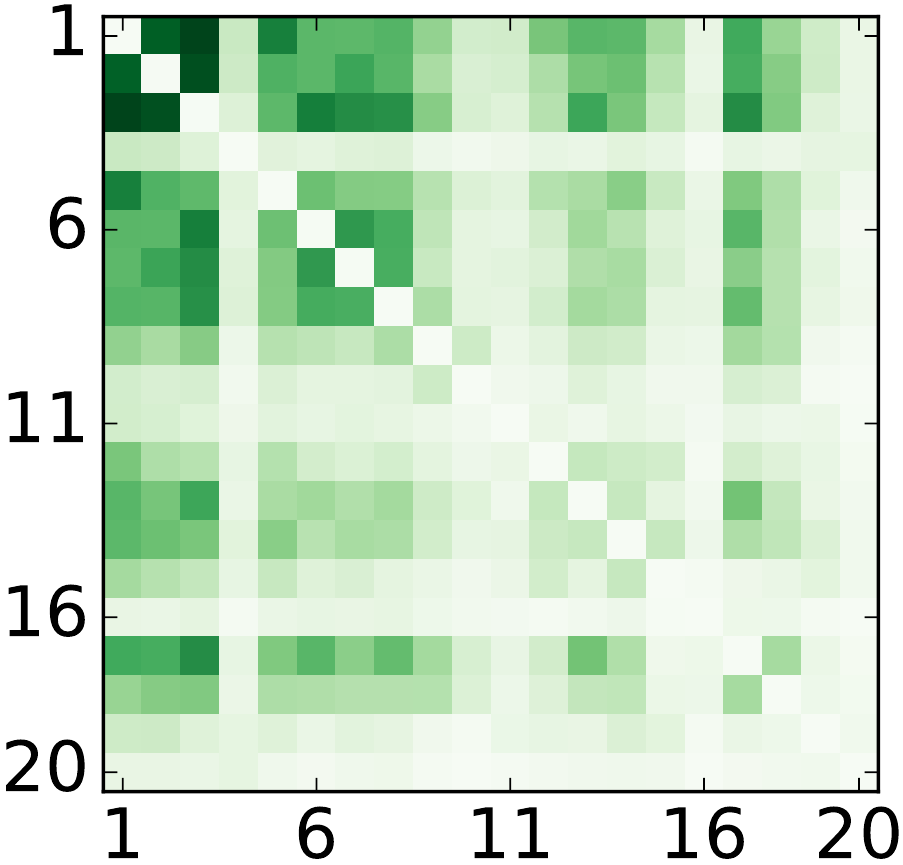}
}
\caption{The co-occurrence frequency of top 20 diagnosis features. Generated data (in green) shares similar pattern as original data (in blue) for the same task. x-axis and y-axis: top 20 features sorted by frequency in the original dataset. Dark (blue/green) parts indicate high co-occurrence frequencies.}
\label{fig:gan-data-cooccur}
\end{figure}


Before testing our semi-supervised prediction models with augmented data, we need to inspect whether the generated data from {\ehrgan} are able to simulate original data well enough, especially for the patient records in the case cohorts. Having the generated data similar to original one is an important precondition to improve our model performance instead of hurting it.
We compared the length and features of the original data ($\mathcal{D}_o$) and generated data $\mathcal{D}_g$ for the two case groups.
As shown in Figure~\ref{fig:gan-data-length}, the generated datasets have similar length distributions to original datasets.
Then we check the most frequency input features from the two datasets.
As shown in Figure~\ref{fig:gan-data-uni-dist}, The generated data keeps similar frequencies for the 100 most frequent features.
Comorbidities (cooccurrences) in patient records are quite useful in clinical prediction tasks.
We select 20 most frequent diagnosis features from these two case cohorts, and show the comorbidity heatmaps in Figure~\ref{fig:gan-data-cooccur}. We can find that both the feature frequencies and the comorbidity clusters are well simulated in our generated datasets.
The list of top 10 diagnosis features for the two cohorts are listed in Table~\ref{tab:top:hf-diagnosis} and \ref{tab:top:d-diagnosis}.
Most of them are common diagnoses in patient records, but with slightly different occurrence frequencies for different cohorts. Our generated models are able to capture the occurrence patterns from different case cohorts and keep those patterns very similar to those in the corresponding original datasets.
These analyses not only verify the quality of our generated data, but also help us get better understandings on
patterns in cohorts for different tasks.

\begin{table}[t]
\centering
\caption{Top 10 most frequent ICD-9 diagnosis codes of heart failure cohort group in the generated data.}
\label{tab:top:hf-diagnosis}
\begin{tabular}{ccll}
\toprule[0.12em]
\begin{tabular}[x]{@{}c@{}}Rank\\in $\mathcal{D}_{g}$\end{tabular} &
\begin{tabular}[x]{@{}c@{}}Rank\\in $\mathcal{D}_{o}$\end{tabular} &
\begin{tabular}[x]{@{}c@{}}ICD-9\\Code\end{tabular} &
Diagnosis Descriptions \\
 \midrule \midrule
1 & 2 & 250.0 & Diabetes mellitus, unspecified \\ \midrule
2 & 1 & 401.1 & Hypertension, benign \\ \midrule
3 & 3 & 427.31 & Atrial fibrillation \\ \midrule
4 & 4 & 401.9 & Hypertension, unspecified \\ \midrule
5 & 5 & 272.4 & Other and unspecified hyperlipidemia \\ \midrule
6 & 6 & 496 & Chronic airway obstruction \\ \midrule
7 & 14 & 585.6 & End stage renal disease \\ \midrule
8 & 7 & 272.0 & Pure hypercholesterolemia \\ \midrule
9 & 9 & 285.9 & Anemia, unspecified \\ \midrule
10 & 8 & 244.9 & Hypothyroidism, unspecified \\
\bottomrule[0.12em]
\end{tabular}
\end{table}

\begin{table}[t]
\centering
\caption{Top 10 most frequent ICD-9 diagnosis codes of diabetes cohort group in the generated data.}
\label{tab:top:d-diagnosis}
\begin{tabular}{ccll}
\toprule[0.12em]
\begin{tabular}[x]{@{}c@{}}Rank\\in $\mathcal{D}_{g}$\end{tabular} &
\begin{tabular}[x]{@{}c@{}}Rank\\in $\mathcal{D}_{o}$\end{tabular} &
\begin{tabular}[x]{@{}c@{}}ICD-9\\Code\end{tabular} &
Diagnosis Descriptions \\
 \midrule \midrule
1 & 1 & 401.1 & Hypertension, benign \\ \midrule
2 & 2 & 401.9 & Hypertension, unspecified \\ \midrule
3 & 3 & 272.4 & Other and unspecified hyperlipidemia \\ \midrule
4 & 4 & 427.31 & Atrial fibrillation \\ \midrule
5 & 6 & 244.9 & Hypothyroidism, unspecified \\ \midrule
6 & 5 & 272.0 & Pure hypercholesterolemia \\ \midrule
7 & 10 & V57.1 & Care involving other physical therapy \\ \midrule
8 & 7 & 285.9 & Anemia unspecified \\ \midrule
9 & 8 & 599.0 & Urinary tract infection, unspecified \\ \midrule
10 & 11 & 496 & Chronic airway obstruction \\
\bottomrule[0.12em]
\end{tabular}
\end{table}


\begin{table*}
\centering
\caption{Performance compassion of different CNN and SSL prediction models on four sub-datasets.}
\label{tab:ssl:res}
\begin{tabular}{lcccc|cccc}
\toprule[0.12em]
& \multicolumn{2}{c}{\task{HF50}} & \multicolumn{2}{c|}{\task{Dia50}} & \multicolumn{2}{c}{\task{HF67}} & \multicolumn{2}{c}{\task{Dia67}}\\ \cmidrule{2-9}
& \textbf{Accuracy} & \textbf{AUROC} & \textbf{Accuracy} & \textbf{AUROC} & \textbf{Accuracy} & \textbf{AUROC} & \textbf{Accuracy} & \textbf{AUROC} \\ \midrule \midrule
\method{CNN-BASIC} & $0.8096$ & $0.8784$ & $0.8990$ & $0.9156$ & $0.8347$ & $0.8953$ & $0.9129$ & $0.9386$ \\ \midrule
\method{CNN-RAND} & $0.7418$ & $0.7856$ & $0.7734$ & $0.8011$ & $0.7788$ & $0.8117$ & $0.7969$ & $0.8486$ \\ \midrule
\method{CNN-FULL} & $0.8631$ & $0.9212$ & $0.9335$ & $0.9528$ & $0.8749$ & $0.9329$ & $0.9486$ & $0.9714$ \\ \midrule
\method{SSL-SMIR} & $0.8207$ & $0.8842$ & $0.9089$ & $0.9197$ & $0.8466$ & $ 0.9102$ & $0.9038$ & $0.9277$ \\ \midrule
\method{SSL-LGC} & $0.8119$ & $0.8767$ & $0.8844$ & $0.9102$ & $0.8325$ & $ 0.9011$ & $0.8815$ & $0.9128$ \\ \midrule
\textbf{\method{SSL-GAN}} & $0.8574$ & $0.9075$ & $0.9135$ & $0.9354$ & $0.8662$ & $ 0.9246$ & $0.9330$ & $0.9563$ \\ \bottomrule[0.12em]
\end{tabular}
\end{table*}

\subsection{Evaluation of the Boosted Model}
To evaluate the performance of the boosted model with semi-supervised learning setting, we conduct extensive experiments on the following six approaches.
\begin{itemize}
\item \method{CNN-BASIC}: The basic model described in Section~\ref{sec:method-basicmodel}, trained only on the training subset;
\item \method{CNN-FULL}: The basic model trained with the same amount of labeled data as \method{SSL-GAN};
\item \method{CNN-RAND}: The basic model trained with the same amount of data as \method{SSL-GAN} with random labels for additional data and true labels for training subset;
\item \method{SSL-SMIR}: Squared-loss mutual information regularization~\cite{icml2013_niu13};
\item \method{LGC}: Semi-supervised learning approach with local and global consistency~\cite{Zhou04learningwith};
\item \method{SSL-GAN}: The proposed method with {\ehrgan} based data augmentation.
\end{itemize}
It is notable that \method{SSL-SMIR} and \method{SSL-LGC} are strong and robust SSL baselines. \method{CNN-FULL}, \method{SSL-SMIR}, and \method{SSL-LGC} are trained with additional samples from a held-off subset. The parameters setting of \method{SSL-SMIR} and \method{SSL-LGC} follows those in the original papers~\cite{Zhou04learningwith,icml2013_niu13} and bag of words feature are used.
We choose the values of $\rho$ and $mu$ in \method{SSL-GAN} with best performance by Section~\ref{sec:exp-params}.
We summarize the classification performance in Table \ref{tab:ssl:res} in different settings with different amounts of labeled data.
For example, \task{HF50} means $50\%$ of the training set of Heart Failure is used, and \task{Dia67} means $2/3$ of the training set of Diabetes is used.
First of all, our model consistently beats \method{CNN-BASIC} and \method{CNN-RAND}.
On \task{HF50}, \method{SSL-GAN} achieves $0.8574$ on accuracy and $0.9075$ on AUROC score, compared with $0.8096$ and $0.8784$ for \method{CNN-BASIC}.
On \task{Dia50} and \task{Dia67}, \method{SSL-GAN} also improves $3\%-4\%$ over the baseline in both measurements.
Due to the messed up label information, the performance of \method{CNN-RAND} is even worse than \method{CNN-BASIC}. Second, compared with the \method{CNN-FULL} method, our model can also achieve comparable results. Measured in AUROC score, \method{SSL-GAN} is about $2\%$ lower than \method{CNN-FULL} on both \task{Dia50} and \task{Dia67} sets.
On \task{HF50} and \task{HF67}, the margins are even smaller.
The two standard SSL methods \method{SSL-SMIR} and \method{SSL-LGC} do not perform very well, only achieving similar performances as \method{CNN-BASIC}, and our method easily beats them.
Overall, these evaluations show the strong boosting power of the proposed \method{SSL-GAN} model.


\begin{table}[ht]
\centering
\caption{AUROC score comparison with different $\rho$.}
\label{tab:pho:auc}
\begin{tabular}{lccc}
\toprule[0.12em]
\textbf{Model} & \task{Heart Failure} & \task{Diabetes} \\
\midrule \midrule
No augmentation ($\mu=0$) & $0.8784$ & $0.9156$ \\ \midrule
\method{SSL-GAN} ($\rho=0$) & $0.8654$ & $0.8754$ \\ \midrule
\method{SSL-GAN} ($\rho=0.001$) & $0.8823$ & $0.9188$ \\ \midrule
\method{SSL-GAN} ($\rho=0.01$) & $0.8911$ & $0.9237$ \\ \midrule
\method{SSL-GAN} ($\rho=0.1$) & $0.9075$ & $0.9354$ \\ \midrule
\method{SSL-GAN} ($\rho=0.2$) & $0.8876$ & $0.9025$ \\ \midrule
\method{SSL-GAN} ($\rho=1$) & $0.7503$ & $0.7603$ \\ \bottomrule[0.12em]
\end{tabular}
\end{table}

\subsection{Selections of Parameters}
\label{sec:exp-params}
We use \task{HF50} and \task{Dia50} to show the effects of values of the two hyper-parameters $\rho$ and $\mu$.

\subsubsection{The effectiveness of \texorpdfstring{$\rho$}{rho}}
In this part, we discuss how the selection of $\rho$ in Equation~\ref{eq:ae} affects the performance. We fix other parameters and vary $\rho$ from $0$ to $1$, and report the AUROC score with different settings in Table~\ref{tab:pho:auc}.
We see that on both datasets, with a properly chosen $\rho$ the generator is able to provide good generations to improve learning. $\rho=0.1$ is an optimal selection for the model (results of $\rho > 0.2$ are no better than $\rho=0.2$ and thus omitted here). On the other hand, with $\rho=0$, which corresponds to sample from an autoencoder, hurts performance. $\rho=1$ completely messes up training as the generated samples are not guaranteed to have the same label as the samples conditioned on. This shows that the transition distribution is able to generate samples that are sufficiently different from training samples to boost the performance.

\begin{figure}[bt]
\centering
\subfigure[Heart Failure]{\includegraphics[width=0.47\columnwidth]{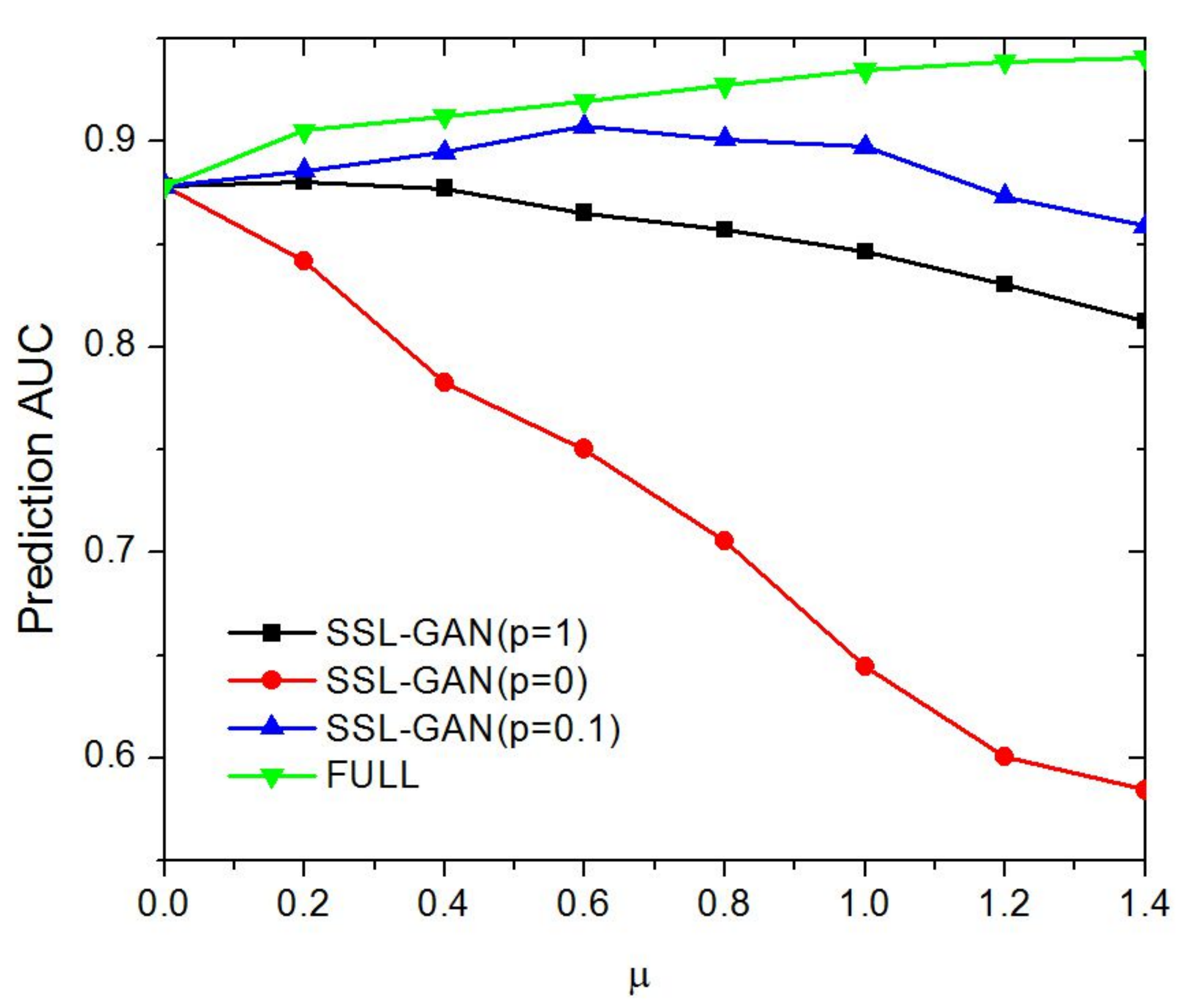}}
\subfigure[Diabetes]{\includegraphics[width=0.47\columnwidth]{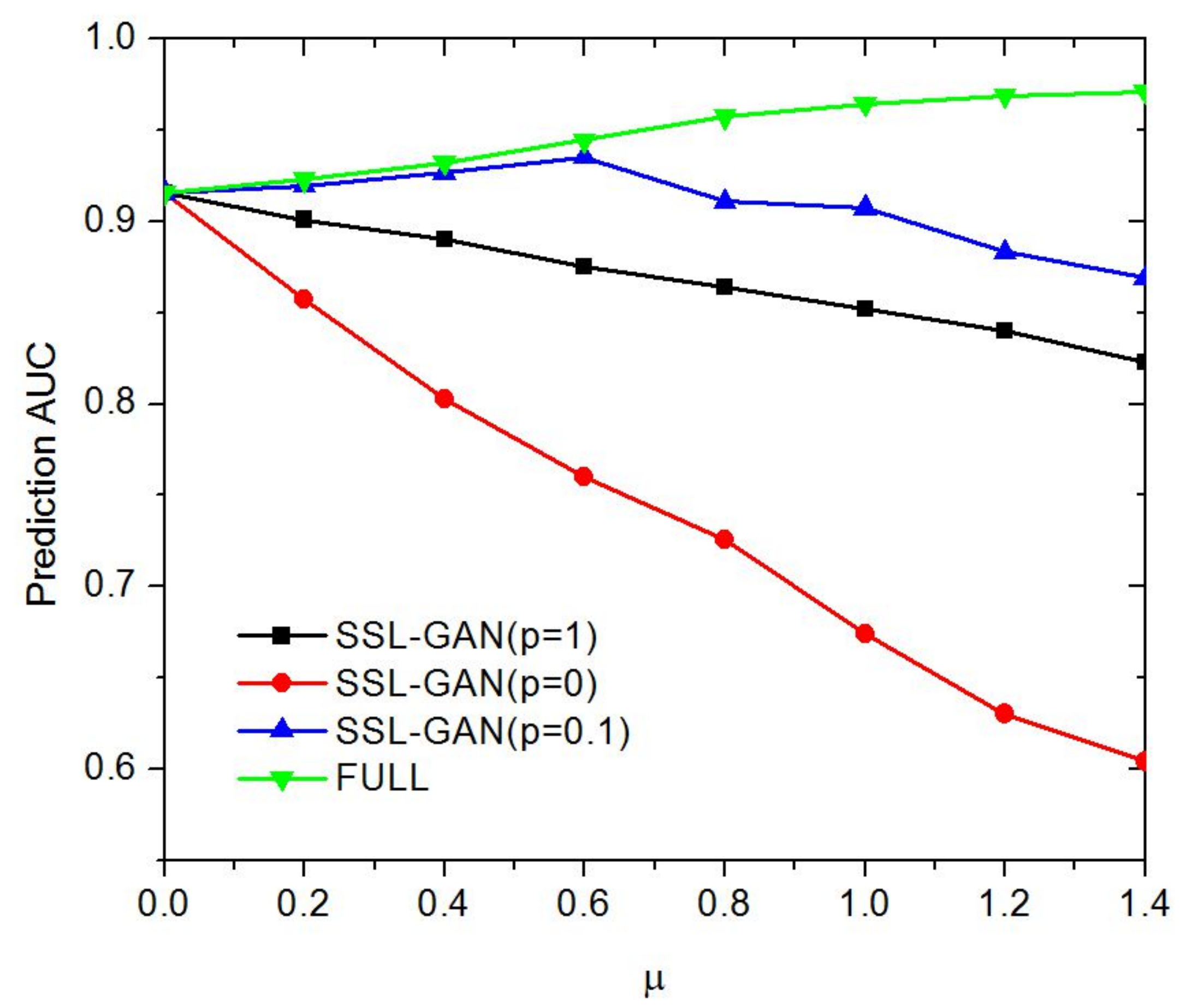}}
\caption{AUROC score comparison with different values of $\mu$.}
\label{fig:mu}
\end{figure}

\subsubsection{The effectiveness of \texorpdfstring{$\mu$}{mu}}
How to optimally utilize the augmented data from GANs to support the supervised learning is an important problem. In our task, this is controlled by the parameter $\mu$ in Equation~\ref{eq:semi}, which leverages the ratio of labeled data and augmented data.
Generally, including more augmented data will help while too many augmented data may even hurt the performance.
With a fixed value of $\rho$, we vary $\mu$ from $0.2$ to $1.4$ and test the prediction performance on two datasets. We also include the setting with fully labeled data (\textit{FULL}), and $\mu$ represents the number of real labeled data used instead of from GANs in this setting.
The prediction AUROC scores of different methods are shown in Figure~\ref{fig:mu}.
It is obvious for the method with fully labeled data that the prediction performance continues improving with $\mu$ increased. For \method{SSL-GAN}, when $\rho=0.1$ (the optimal setting), we can see it achieved the best performance when $\mu=0.6$. After that point, the performance decreased a little, which indicates that more augmented data can not help further. For the setting with $\rho=1$ and $\rho=0$, the prediction power continues falling as including more augmented data is harmful for both cases. Similar trend are also observed under the measure of accuracy.

\section{Conclusion} \label{conclusion}
In this paper, we focus on exploiting deep learning technique and its applications in healthcare. We first present {\ehrgan}, a generation model via adversarial training, and discuss several techniques for learning such a model for EHR data. We demonstrate that the proposed model can produce realistic data samples by mimicking the input real data, and the learned latent representation space can continuously encode plausible samples. To boost risk prediction performance, we utilize the learned model to perform data augmentation by semi-supervised learning. Experimental results on two datasets show that the proposed model improves the generalization power and the prediction performance compared with strong baselines.

In future work, we would like to perform more comprehensive quantitative comparisons from a clinical view with the help of domain experts.
This includes analyzing the important clinical patterns conditioning on disease, the concurrence of several diagnoses, and the correlations among other key medical features.
The proposed generation model can also be improved with better clinical interpretation or structure information and achieve more compelling results. We may also try to incorporate state-of-art GANs \cite{DBLP:journals/corr/LiCCYP17} into our framework. Furthermore, other boosted learning techniques with GANs, such as training a joint learning model, and sharing latent representation space between the networks, will be evaluated on EHR data. Finally, the proposed framework can be naturally extended to other healthcare applications, such as readmission prediction and representation learning. 

\bibliographystyle{IEEEtran}
\bibliography{references}

\end{document}